\def\year{2017}\relax
\documentclass[letterpaper]{article} %DO NOT CHANGE THIS
\usepackage{aaai17}  %Required
\usepackage{times}  %Required
\usepackage{helvet}  %Required
\usepackage{courier}  %Required
\usepackage{url}  %Required
\usepackage{graphicx}  %Required
\usepackage[ruled]{algorithm2e}
\usepackage{multirow}
\usepackage{subcaption}
\graphicspath{ {images/} }
\begin{document}
% The file aaai.sty is the style file for AAAI Press 
% proceedings, working notes, and technical reports.
%

\title{Toward Crowd-Sensitive Path Planning}
\author{Anoop Aroor\textsuperscript{1} and Susan L. Epstein\textsuperscript{1,2}\\
The Graduate Center\textsuperscript{1} and Hunter College\textsuperscript{2}  of The City University of New York\\
New York, NY 10065\\
aaroor@gradcenter.cuny.edu, susan.epstein@hunter.cuny.edu
}
\maketitle
\begin{abstract}
If a robot can predict crowds in parts of its environment that are inaccessible to its sensors, then it can plan to avoid them. This paper proposes a fast, online algorithm that learns average crowd densities in different areas. It also describes how these densities can be incorporated into existing navigation architectures. In simulation across multiple challenging crowd scenarios, the robot reaches its target faster, travels less, and risks fewer collisions than if it were to plan with the traditional A* algorithm. 
\end{abstract}

\section{Introduction}

\noindent Robots increasingly serve in environments shared with people, as museum tour guides \cite{thrun1999minerva}, telepresence robots \cite{Tsui:2011:EUC:1957656.1957664}, and assistants in hospitals and offices \cite{veloso2015cobots}. When these environments are crowded, \textit{autonomous navigation} (the ability to move about without human intervention) becomes challenging. Because the robot can detect people only in its vicinity, it is restricted to \emph{local} rather than \emph{global} crowd data. Moreover, because crowds move, path planners that assume a static world may generate infeasible or inefficient plans. In addition to travel distance and time, in the presence of crowds, other criteria become important, including safety, comfort, and social mores \cite{kruse2013human}.  The thesis of our work is that a robot can meet the challenges of a crowded environment when it learns to predict global crowd behavior and to plan with that knowledge. We call this \emph{crowd-sensitive planning}.

Ideally, a robot familiar with its environment should be able to predict crowded areas and simply avoid them. This would reduce travel time, travel distance, and the likelihood of collision. A robot that could detect crowds only within its sensor range might generate the plan shown in Figure 1 as a solid line. As a result, the robot would repeatedly correct this plan (\emph{plan repair}) or make a new one (\emph{replan}), and pause often until it had an opportunity to move. In contrast, a crowd-sensitive plan, shown as a dashed line in Figure 1, might entirely avoid areas likely to be crowded.

\begin{figure}
\includegraphics[width=0.45\textwidth]{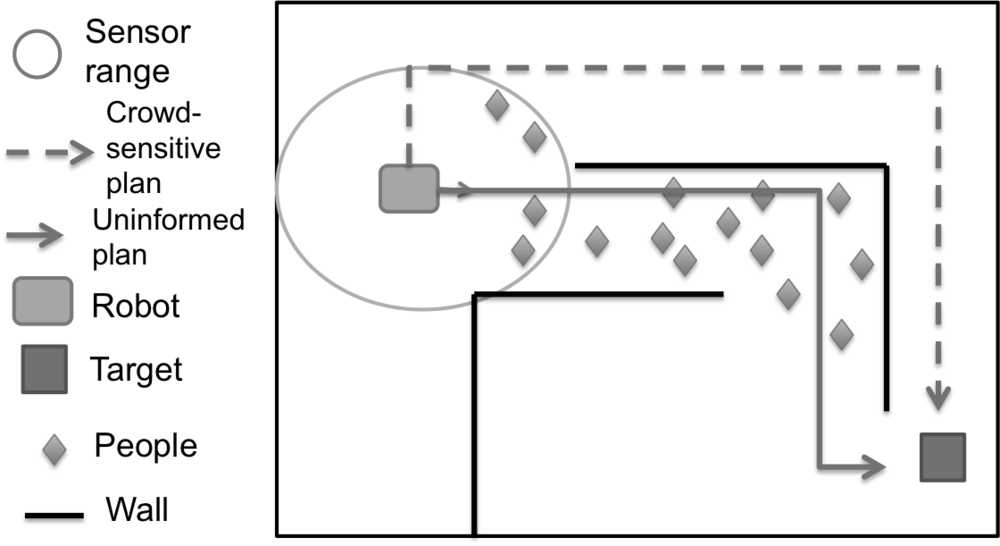}
\caption{Example to demonstrate the usefulness of a crowd-sensitive plan}
\end{figure}

A \textit{crowd density map} is an observed, cumulative, global record of the crowd in an environment. A robot can use a crowd density map  to predict where people are likely to appear. Our approach learns a crowd density map for a two-dimensional space as a robot moves in it from one \emph{location} (coordinates (\emph{x,y}) with respect to an allocentric origin) to another. 

The contributions of this paper are threefold. First, it introduces a fast, online algorithm that learns a crowd density map from only local sensor information, with no assumptions about the size or behavior of the crowd. Second, it describes how crowd density maps can be incorporated into existing navigation architectures to generate crowd-sensitive plans. Finally, it demonstrates empirically, through extensive simulation, the benefits of crowd-sensitive planning. In particular, it shows statistically significant improvements in both efficiency and safety over a traditional A* planner. 

The next section provides background and describes related work. Subsequent sections list the assumptions behind our approach, describe how to learn a crowd density map, and explain how to generate crowd-sensitive plans to improve both navigation and safety. These are followed by the experimental design, results, and a discussion.

\section{Background and Related Work}

Early research in autonomous navigation for crowded environments simply detected local obstacles and then sought to avoid them; it did not predict their future motion. One such approach used a dynamic window \cite{fox1997dynamic}. Search for commands to control the robot was done in \emph{velocity space}, the set of all velocities (vectors that indicate the speed and direction of motion) that can be sent to the robot as motion commands. Velocity space was pruned to reflect the physical constraints of the particular robot. Among the remaining velocities, one that maximized the objective function was chosen. Such approaches, that either replan or make local changes to a plan only after they detect local obstacles, can be severely hampered by crowds. 

The robot's \emph{actuators} (which convert electrical energy into physical movement) and \emph{sensors} (which convert measurements of the environment into electrical signals) are imperfect, noisy, and can deteriorate over time. As a result, the robot might not detect a person, or might move too much or too little. Moreover, people, particularly children, do not always maintain a safe distance from the robot \cite{Nomura:2015:WCA:2701973.2701977}. One way to address such uncertainty is to specify a large minimum distance to people that is considered safe. While this might reduce the risk of collision, it also increases navigation time because the robot must wait longer, move more slowly, or take longer paths to avoid people. In contrast, our approach creates plans that improve safety without sacrificing travel time or distance.

To improve collision avoidance, many methods have made local, short-term predictions about the future trajectories of moving obstacles with, for example, Gaussian processes \cite{trautman2010unfreezing} or neural networks \cite{alahi7780479}. Biomechanical turn indicators have been used to predict short-term trajectories and then plan around them \cite{unhelkar2015human}. Another planning approach was cooperative, where people and robots gave way for one another \cite{trautman2013robot}. A local path planner learned cost functions on data from human experts who controlled the robot \cite{kim2016socially}. Yet another approach used pedestrian trajectory datasets to learn a model that jointly predicted the trajectories of both a robot and nearby pedestrians, and then generated socially compliant paths \cite{kretzschmar2016socially}. Although these approaches improved collision avoidance and moved more safely near people, their robots could still generate global path plans through crowded areas, because they predict trajectories only when the robot senses pedestrians in its vicinity. Our work, in contrast, addresses the complementary problem of how to learn a global crowd behavior model in a given environment and use it to improve global path planning. 

Many other approaches have made global, long-term predictions about the behavior of a crowd, and adapted their navigation behavior accordingly. One approach treated a single trajectory as a Markov decision process, and learned a distribution over trajectories \cite{Dey5354147}. With inverse reinforcement learning, it learned the reward function that best fit the trajectories, and used it to predict new ones. Another trajectory predictor created human ego graphs that could be queried efficiently for future trajectories  \cite{Chung2010AMR}. A model based on Gaussian processes learned the global distribution of the crowd, and then used inverse reinforcement learning to make the robot's behavior more human-like  \cite{Henry5509772}. End-to-end pedestrian trajectory data from the simulator is used to calculate initial estimates of mean crowd densities in unobserved areas of the grid, and then local sensor data are used to update these densities. Such work, however, requires a dataset of complete pedestrian trajectories, which is unavailable to a robot that can sense only locally. Our work, in contrast, assumes only local sensor data. It does not require trajectory data or separate phases for learning and testing. Instead, it learns online as it completes its task.

\begin{figure}[b]
\centering
\includegraphics[width=0.45\textwidth]{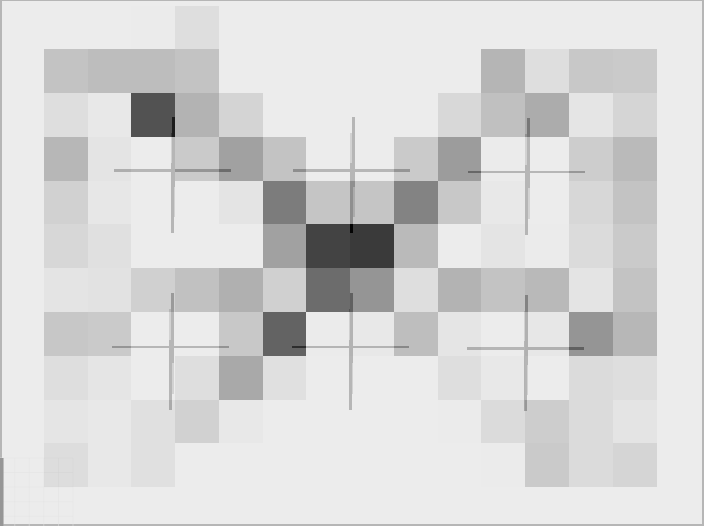}
\caption{A crowd density map}
\end{figure}

\section{Approach}
This work makes several assumptions, as follows. The robot has a two-dimensional map of the static features in its environment, and has laser range sensors mounted at a uniform level near the floor. The robot localizes itself from this information, that is, at any point the robot knows its \emph{pose} (location and orientation) with respect to an allocentric coordinate system. The robot also detects only \emph{local crowd data}, the location and direction of motion of each person within its sensor range \cite{People7139259}.

The robot's \emph{task} is to begin from an initial pose and move through a crowded environment to visit a sequence of locations (\emph{targets}). To do so, the robot executes a sequence of control cycles. Each control cycle is a four-phase sense-learn-decide-act loop: the robot senses its environment, learns from its sensed data, selects an action with a decision-making algorithm, and then executes it. Possible actions from which to choose are a discretized set of forward moves and rotations. The execution phase sends the chosen action to the actuators to move the robot. 

\subsection{Learning crowd density maps}
To predict where people are likely to appear in the environment, the robot learns online, as it travels. Local crowd data from the sense phase is forwarded to a learning module that updates the values in the crowd density map. The robot can use this observed, cumulative, global record of the crowd to predict where crowds are likely to obstruct its passage. 

Our work represents a crowd density map as an $r \times c$ grid superimposed on the two-dimensional footprint of the environment. Each cell has a \emph{density}, the running average of the number of people the robot has observed within that cell over time.
When the robot enters the sense-learn-decide-act loop at time $t$, it converts the range data from its sensors into local crowd data $L_t$,  the locations of all people detected within sensory range. For $p$ people detected at time $t$ 

\[ L_t = \{(x,y)_t^1,...(x,y)_t^p\}\]
For each grid cell in row $i$ and column $j$, the map maintains three values that summarize the robot's experience: $k_{ij}$, the number of times the robot has collected information about the cell; $t_{ij}$, the number of people the robot has detected in the cell; and the \emph{crowd density} $d_{ij} = t_{ij} / k_{ij}$.

Algorithm 1 is a fast update algorithm that performs online computation of the crowd density map. For each $L_t$, the first loop produces a temporary count $curr_{ij}$ of the number of people currently detected in the $ij$th cell. The second loop iterates over the crowd density map to update $t_{ij},  k_{ij}$ and $d_{ij}$. As it updates, it also ages the values of the previous observations $t_{ij}$ and $k_{ij}$ with \emph{discount factor} $\alpha \in (0,1]$. It increments $k_{ij}$ by 1, increments $t_{ij}$ by $curr_{ij}$, and recomputes $d_{ij}$ based on the new values. Algorithm 1 has complexity $O(r \cdot c + P)$, linear in the number of grid cells and the maximum number of people $P$ permitted in the environment. An example of a learned crowd density map is shown in Figure 2, where darker cells have higher $d_{ij}$ values.

\begin{algorithm}
\caption{Crowd density map update}
\SetAlgoLined
\KwIn{$t_{ij},k_{ij},L_t$}
\KwOut{ $d_{ij}$ }
 $curr_{ij} = 0, \forall (i,j) \in \{1,..r\} \times \{1,..c\}$\;
 \For{$(x,y) \in L_t$}{
  $(i,j) = convertToGridIndex(x,y) $\;
  $curr_{ij} = curr_{ij} + 1$
 }
 \For{$(i,j) \in \{1,..r\} \times \{1,..c\}$}{
  $t_{ij} = (t_{ij} * \alpha) + curr_{ij}$\;
  $k_{ij} = k_{ij} * \alpha$\;
  \If{$isGridVisible(i,j)$}{
    $k_{ij} = k_{ij} + 1$
  }
  $d_{ij} = t_{ij}/k_{ij}$
 }
\end{algorithm}

\subsection{SemaFORR-based navigation}
In this work, the robot's decision algorithm is \emph{SemaFORR}, a controller for autonomous navigation \cite{Epstein:2015:LSM:2965680.2965705}. SemaFORR is implemented in ROS, the state-of-the-art Robot Operating System \cite{quigley2009ros}. To do crowd-sensitive planning, we introduce a crowd density map into SemaFORR.

\begin{figure}
\includegraphics[width=0.45\textwidth]{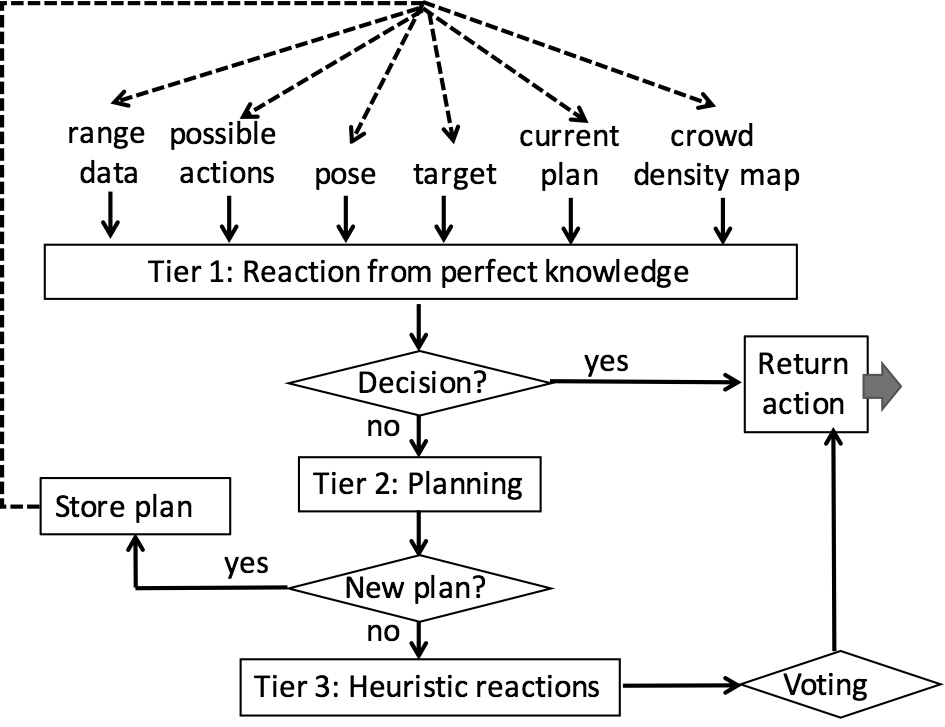}
\caption{SemaFORR's decision cycle. %Dashed lines indicate information flow; solid lines indicate the control flow. 
The crowd density map is the new addition described here}
\end{figure}

\begin{figure*}[t]
\centering
\begin{subfigure}[]{.32\linewidth}
    \centering
    \includegraphics[width = \linewidth]{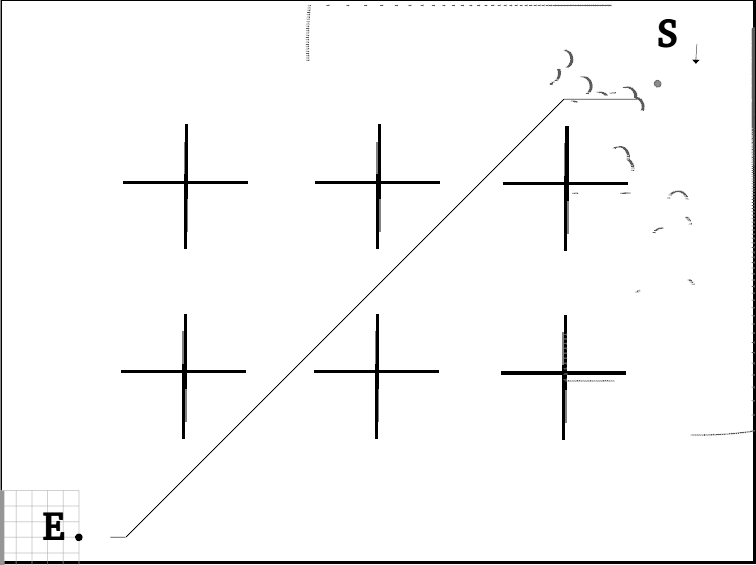}
    \caption{}
\end{subfigure}
\hfill
\begin{subfigure}[]{.32\linewidth}
    \centering
    \includegraphics[width = \linewidth]{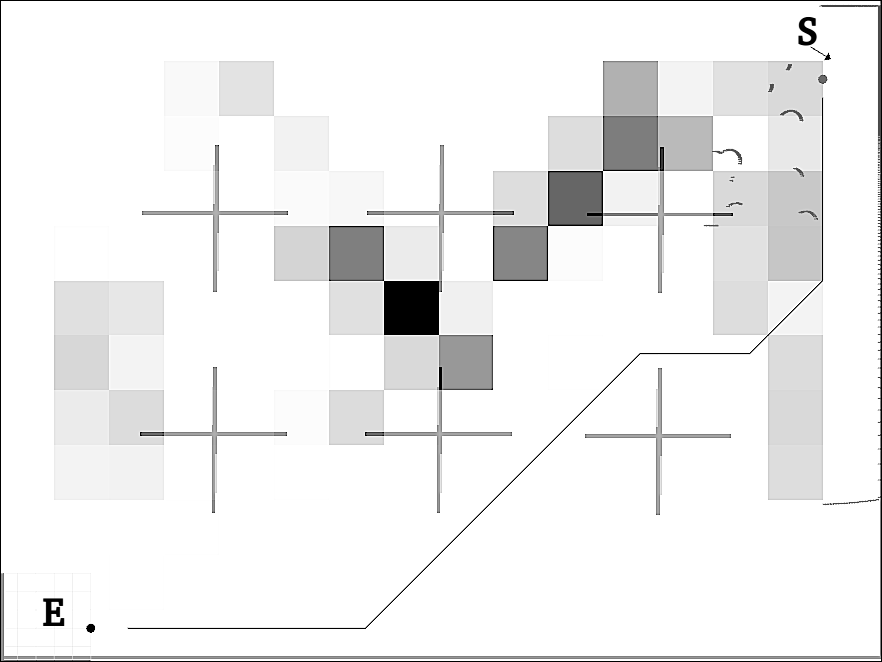}
    \caption{}
\end{subfigure}
\hfill
\begin{subfigure}[]{.32\linewidth}
    \centering
    \includegraphics[width = \linewidth]{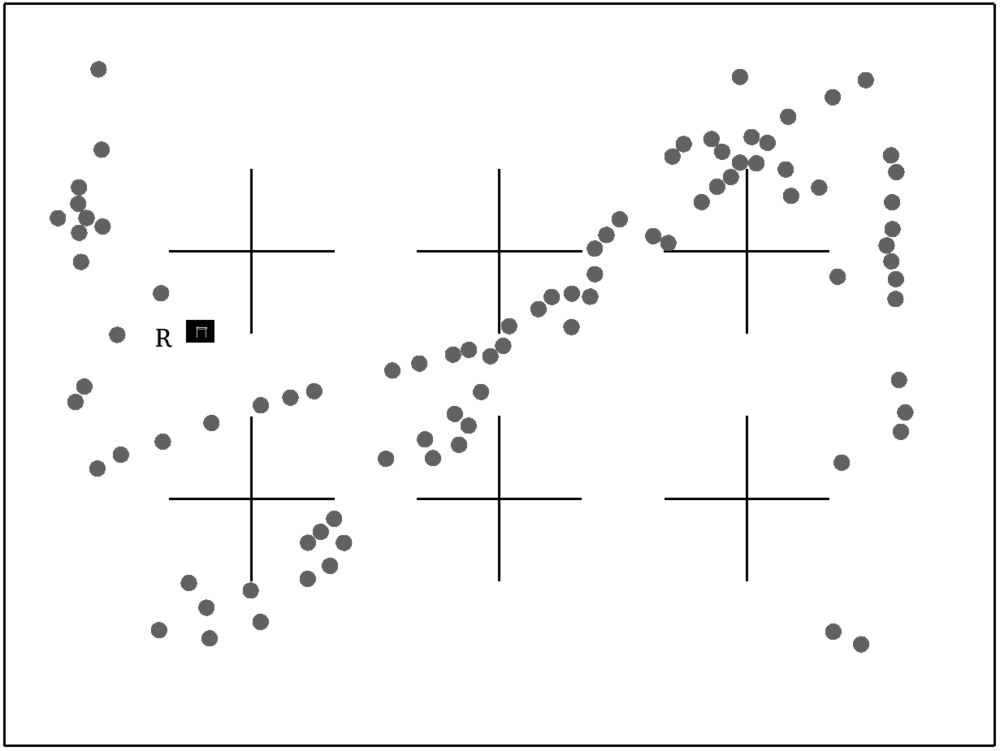}
    \caption{}
\end{subfigure}
\caption{The map of a simple, cubicle-like, 48m $\times$ 36m office environment with interior walls as heavy lines. (a) Traditional A* plan for the robot from start S to end E (b) Crowd density map and CSA* plan for the same task (c) Zig-zag crowd behavior for 90 people in MengeROS, with a black box that indicates the robot's initial position in the experiments reported here} 
\end{figure*}

SemaFORR is a cognitively-based hybrid architecture that involves both reactive and deliberative reasoning. The deliberative reasoning component generates a plan that is a sequence of intermediate locations (\textit{waypoints}) on the way to the target. As in Figure 3, SemaFORR's input includes the actions available to the robot, its pose and  current target, the current laser scan data, and the crowd density map. Because SemaFORR is based on the FORR cognitive architecture \cite{COGS:COGS275}, it uses a combination of heuristic procedures called \emph{Advisors} to choose an action. SemaFORR's Advisors form a three-tier hierarchy. 

\emph{Tier-1} Advisors are reactive decision-making rules that assume perfect knowledge. As a result, they are fast and correct. Each Advisor can either choose an action to execute or eliminate actions from further consideration. \textsc{Victory} is a tier-1 Advisor; it chooses the action that gets the robot closest to the target when it is within sensory range and no obstacles block the robot's path. If there is such an action, it  is executed and the cycle ends. Otherwise, \textsc{AvoidObstacles}, another tier-1 Advisor; uses the laser range scan data to eliminate actions that would cause a collision or bring it too close to static or dynamic obstacles. If only one action remains, it is returned. Finally, if the robot has a plan to reach the target, at least one unvisited waypoint is within sensory range, and no obstacles block the robot's path to it, \textsc{Enforcer} selects the action that best approaches the waypoint closest to the target. If there is such an action, it is returned. Otherwise, SemaFORR proceeds to tier 2.

\emph{Tier-2} Advisors are deliberative planners. If there is a current plan, tier 2 forwards the remaining actions to tier 3. Otherwise, in this implementation, there are two planners, only one of which is active in any given experiment: A* and \emph{CSA*} (Crowd-Sensitive A*, described below). If there is no current plan, the planner creates one, SemaFORR stores it, and the cycle ends.  

\emph{Tier-3} Advisors make heuristic recommendations that may or may not be correct, but are based on a single rationale. Given the current plan, tier-3 Advisors treat the next waypoint as the target. For example, \textsc{Greedy} prefers actions that move the robot closer to that waypoint, and \textsc{Explorer} prefers actions that move the robot away from previously visited areas. Each tier-3 Advisor expresses its preferences as a numerical value for each of the actions that remained after tier 1.  A voting mechanism aggregates the preferences of all tier-3 Advisors and returns the most preferred action. After the execution of a returned action, the decision cycle ends, and the sense-learn-decide-act loop resumes with updated input. Further details on SemaFORR are available in  \cite{Epstein:2015:LSM:2965680.2965705}.

\begin{figure}[b]
\centering
\includegraphics[width=0.45\textwidth]{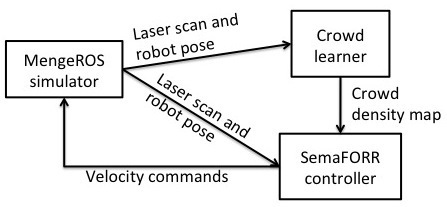}
\caption{ROS node interactions}
\end{figure}

\subsection{Avoiding crowded areas}
A* is the traditional, optimal, search-based path planner for static maps \cite{astar4082128}. In this implementation, A* reasons over a graph based on a discretized version of the world, the cells of a grid superimposed over its footprint. Each node in the graph represents a grid cell, and is connected to at most eight cells that adjoin it in the grid. (There will be fewer than eight if the cell lies on the border of the grid or a wall intervenes.) The weight of edge $e_{mn}$ that connects nodes $m$ and $n$ is the Euclidean distance between their centers. 

To avoid crowded areas, CSA* adapts A* to plan with a crowd density map.  Given a new target, CSA* queries the learning module for the current crowd density map, and uses it to update the edge weights of the A* graph. This effectively imposes a penalty on travel to crowded areas, so that the resultant CSA* plan moves through less-crowded areas. At the beginning of a task, when the robot receives its first target, CSA* generates a plan based only on travel distance, similar to the solid line in Figure 1. As the robot navigates through its environment, it updates the crowd density map, and CSA* plans increasingly avoid crowded areas, similar to the dashed line in Figure 1.

Before planning, tier-2 retrieves the latest crowd density map. The crowd density variables $d_{ij}$ are normalized over the full grid to range between 0 (least crowded) and 1 (most crowded). The normalized crowd density variables $D_{ij}$ are then used to compute new edge weights for the graph, as follows:

\[ e^{new}_{mn} = e^{old}_{mn} * (D_{ij} + 1) * (D_{kl} + 1)\]
where $(i,j)$ and $(k,l)$ are the grid-cell indices of nodes $m$ and $n$ respectively.
    
When crowd density values at both nodes incident on an edge are 0, the edge weight is unchanged. When they are both 1, the new edge weight is 4 times the old edge weight. Edge-weight increments are a strong influence away from crowded areas. For a sample task that moves the robot from S to E,  Figure 4  compares a traditional A* plan in (a) to a  CSA* plan informed by the crowd density map in (b). A* ignores the likely delays from the crowd. The crowd-sensitive plan is longer but passes through less crowded areas.

\section{Implementation}
Our system is implemented as three interacting ROS nodes, as shown in Figure 5.

\textbf{MengeROS.} To simulate crowding and the robot in a single environment, we use \emph{MengeROS} \cite{mengeROS}, a ROS extension of the open-source crowd simulator \emph{Menge} \cite{curtis2016menge}. MengeROS requires a map of the environment, a robot, and crowd specifications.

\textbf{SemaFORR.} The SemaFORR controller node is initialized with a sequence of target points. It receives the simulated robot's position and laser scan data from the MengeROS node, and returns  to the MengeROS node the actuator commands chosen by SemaFORR. The MengeROS node then simulates that action on the robot.

\textbf{Crowd Learner.} The Crowd Learner is a standalone ROS node; it too receives the robot's position and laser scan data as messages from MengeROS. The Crowd Learner uses Algorithm 1 to update the crowd density map, and forwards the revised crowd density map to the SemaFORR node. This modular implementation is important because it allows the learner node to be used with any other ROS-compatible simulator or any other ROS-compatible robot controller.

\section{Experimental Design}
In these experiments, MengeROS simulates the footprint and sensor readings of Freight, an affordable standard platform for mobile service robots \cite{wise2016fetch}. Freight has a 2-D laser range scanner with a range of 25m and a 220$^\circ$ field of view, a 15Hz update rate, and an angular resolution of $\frac{1}{3}^\circ$. Each experiment sets a task for Freight, controlled by SemaFORR, in an environment that it shares with a crowd controlled by MengeROS. 

MengeROS defines \emph{crowd behavior} by the initial positions of its members, a state transition diagram for each person that specifies how to select her next target, and a uniform decision mechanism to select moves. These experiments use two crowd behaviors: random and zig-zag. Both begin with the robot in the position shown in Figure 4(c), and the crowd in the lower left corner.
Under \emph{random crowd behavior}, each person's next destination is chosen randomly from among six prespecified locations (one in each corner of the map and two evenly spaced in the center), and each person follows an A* plan to her own target. Such a crowd moves as if it were in a complex subway station with multiple destinations. Under \emph{zig-zag crowd behavior}, the crowd moves within the pattern shown  in Figure 4(c). Each person chooses one of three random points in the upper left corner, moves to it with an A* plan, then chooses one of three random points in the lower left corner and moves to it under A*. This process continues, so that the crowd loops though upper left, lower left, upper right, lower right, and back to the upper left corner again, as if it were parading in a figure eight with some slight internal variation.  Such a crowd presents a less even challenge that intensifies along its route.

MengeROS also requires a uniform collision avoidance strategy that all its members use to avoid one another and the robot. These experiments use two collision avoidance strategies: ORCA \cite{vandenBerg2011} and PedVO \cite{curtis2014pedestrian}. Both strategies are based on velocity obstacles. The velocity obstacle (\emph{VO}) of a person is the set of all velocity vectors that will result in collision. Collision-free motion requires that every agent have a velocity vector outside its VO. ORCA has each agent address this problem equally to produce an optimal solution. PedVO adapts ORCA to behave more similarly to people; it introduces such human behaviors as aggression, social priority, authority, and right of way. 

We use 12 possible MengeROS \emph{crowd scenarios}, each defined by its crowd size (here, 30, 60, or 90 people), collision avoidance strategy (PedVO or ORCA), and crowd behavior (random or zig-zag). An \textit{experiment configuration} specifies a crowd scenario, one of two target sets for the robot (A or B, each a list of 15 randomly chosen locations), and whether the Crowd Learner node is on or off. When it is off, tier 2 uses the traditional A* planner that minimizes distance; when it is on, tier 2 computes crowd-sensitive plans with CSA*. Thus there are $12 \cdot 2 \cdot 2 =  48$ experiment configurations in all. For all experiments reported here with CSA*, the discount factor $\alpha$ was set to 1.

Given an experiment configuration, an \textit{experiment} executes decisions from SemaFORR on the simulated robot until the robot \emph{reaches} (comes within $\epsilon$ of) each of its targets in the prespecified order. Each configuration was executed 5 times on an 8-core, 1.2 GHz workstation. Evaluation metrics are the total time the robot took to reach the targets, the total distance it travelled, the number of the robot's \emph{risky actions} (ones that placed it less than 0.5 meters from a person or a wall), and  \textit{clearance}, the average minimum distance the robot maintained from all obstacles. Clearance and risky actions recognize important concerns that arise when people crowd an environment.

\begin{table}[b]
\centering
\label{my-label}
\begin{tabular}{|l|r|r|r|}
\hline
               & \multicolumn{1}{c|}{\textbf{A*}} & \multicolumn{1}{c|}{\textbf{CSA*}} & \multicolumn{1}{c|}{\textbf{\% change}} \\ \hline
time (sec.)    & 905.51                           & 618.03                             & -31.7\%***                              \\ \hline
distance (m.)  & 843.81                           & 624.38                             & -26.0\%***                              \\ \hline
clearance (m.) & 1.82                             & 2.02                               & 11.50\%**                                \\ \hline
risky actions  & 366.06                           & 169.50                             & -53.7\%***                              \\ \hline
\end{tabular}
\caption{Impact of CSA* vs A* on performance, with improvements. *** denotes $p < .001$; ** denotes $p = .05$.} 
\end{table}

%\begin{table}[]
%\centering
%\label{my-label}
%\begin{tabular}{|l|l|l|l|l|}
%\hline
%              & \textbf{30} & \textbf{60} & \textbf{90} & \textbf{\textit{p}} \\ \hline
%time (sec.)         & 596.22      & 741.66      & 947.43      & ***             \\ \hline
%distance (m.)     & 609.27      & 733.69      & 859.33      & ***             \\ \hline
%clearance (m.)   & 2.02        & 2.03        & 1.70        & ***             \\ \hline
%risky actions & 166.53      & 204.59      & 432.23      & ***             \\ \hline
%\end{tabular}
%\caption{Impact of crowd size on performance. Asterisks denote significance level}
%\end{table}

\begin{figure*}[]
\includegraphics[width=0.24\textwidth]{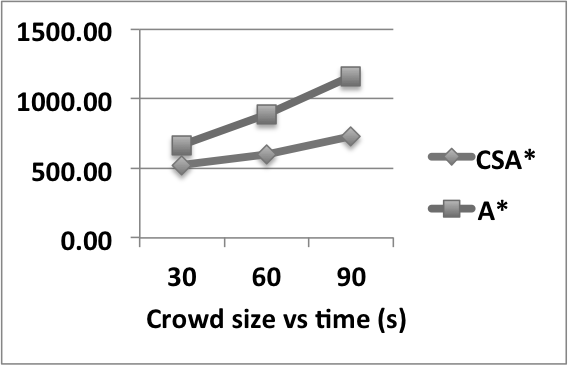}
\includegraphics[width=0.24\textwidth]{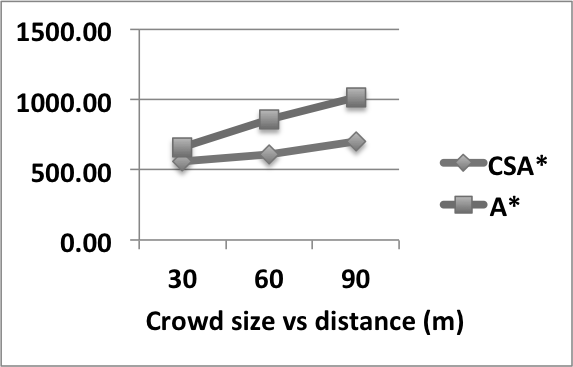}
\includegraphics[width=0.24\textwidth]{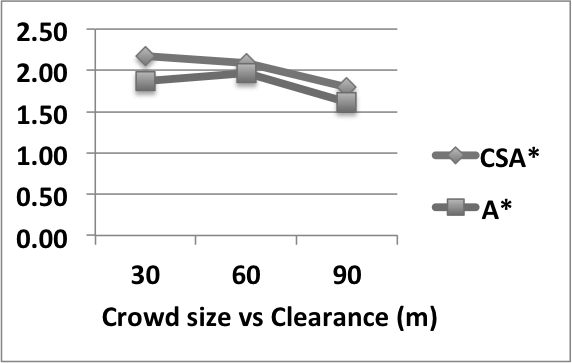}
\includegraphics[width=0.24\textwidth]{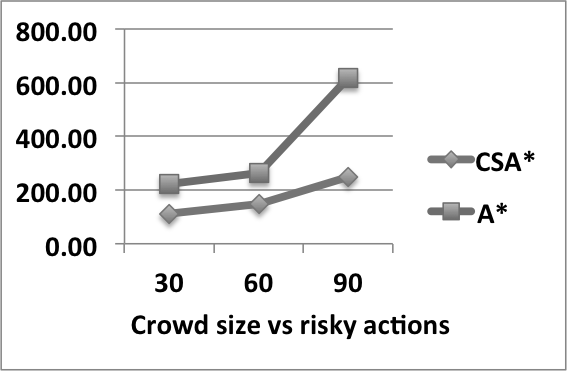}
\caption{Total time, total distance, clearance and risky actions averaged over all CSA* runs vs all A* runs}
\end{figure*}

\section{Results}

Crowd-sensitive planning (Table 1) had a statistically significant effect on every metric, that is, with crowd-sensitive planning, time was faster ($t = 8.57, p < .001$), distance was shorter ($t = 7.99, p < .001$), clearance was larger ($t = 5.28, p < .001$), and risky actions were fewer ($t = 5.65, p < .001$). The effect of crowd-sensitive planning is particularly noteworthy because the space was relatively sparsely populated; even 90 people in a 48m $\times$ 36m space amount to only about 1.3 people in a 5m $\times$ 5m room. Similarly, the crowd size had a statistically significant effect on every metric. As one would expect, the presence of more people delays the robot ($R^2 = 0.24, p < .001$), and makes it travel further ($R^2 = 0.18, p < .001$) and come closer to people and to walls ($R^2 = 0.16, p < .001$), so that travel is generally less safe ($R^2 = 0.14, p < .001$). 

The randomly-generated target lists also had an effect on the performance. Inspection revealed that target set A was an easier task than B, that is, without crowds, the optimal distance to visit A is simply less than the optimal distance to visit B. Thus travel to the A set was faster ($t = 2.12, p = .05$) and covered less distance ($t = 2.28, p = .05$). There were also significant effects from crowd flow. Under zig-zag the robot was faster ($t = 2.28, p = .05$), maintained greater clearance ($t = 6.27, p < .001$), and took fewer risky actions ($t = 3.94, p <  .001$) than it did under the less predictable random crowd flow. The collision avoidance method, however, had no significant effects. This was because both ORCA and PedVO are local methods and therefore do not impact the overall crowd distribution in the environment. 

A multi-factor ANOVA determines the effects of independent variables on a continuous dependent variable. The independent variables were crowd-sensitive planning (CSA* or A*), collision avoidance strategy (ORCA or PedVO), the robot's target list (A or B), the crowd size (30, 60 or 90), and the crowd behavior (random or zig-zag). The dependent variable was each of the metrics in turn: total time to complete the target set, distance traveled, clearance, or number of risky actions. Analysis showed significant effects on time (F(6, 233) = 50.32, $p < .001$), distance (F(6, 233) = 35.96, $p < .001$), clearance (F(6, 233) = 31.27, $p < .001$) and risky actions (F(6, 233) = 20.08, $p < .001$).  

An interaction term was also included in the ANOVA to detect any interaction between crowd size and CSA*. The results, shown in Figure 6, indicate significant interaction effects between crowd size and CSA* for time  (F(6, 233) = 52.10, $p < .001$) and distance (F(6, 233) = 35.58, $p < .001$). This implies that as the environment becomes more crowded CSA* will provide larger improvements in time and distance than A*.

To confirm the magnitude of the changes in Table 1, we also measured effect size with Cohen's $d$ statistic \cite{cohen1962statistical}. This statistic is typically used on social or biological data. It identifies a clear effect if $d$ exceeds the threshold 0.2 and a ``large effect" if it exceeds 0.8. When we paired CSA* and A* experiments and compared them, CSA* had an effect on 90\% of them (threshold 0.2) and a large effect on 75\%. CSA* performed as much as $55\%$ faster than A* (target set B, PedVO, crowd size 90), traveled $51\%$ less far (target set B, PedVO, crowd size 90), had  a $50\%$ higher clearance (target set A, ORCA, crowd size 90) and took $78\%$ fewer risky actions (target set B, ORCA, crowd size 60). Moreover, only 9\% of such comparisons showed any deterioration in performance.

\section{Discussion}
Crowd-sensitive planning is current work. Experiments underway explore the breadth of its effect in other (larger, more complex) maps, with crowd flow behaviors other than random and zig-zag, and with crowds whose destinations change over time (and thereby make it more difficult to predict their presence). We expect further improvements as we learn not only about the number of people likely to be in a grid cell, but also about the directions in which they move. 

In these experiments, a robot that used A* repaired its plan when obstacles interfered with it. We will also explore comparisons to  planners that replan and benefit from previous planning knowledge, such as D* Lite and MPGAA* \cite{Koenig:2002:DLI:777092.777167,Hernandez:2015:RPF:2887007.2887168}.

The granularity (i.e., cell dimension) of the crowd density map here was 3 meters, determined by inspection. A finer crowd density map would provide more detail but require more data to produce an accurate representation and increase the computation time, while a coarser grid would provide less specific guidance. Moving objects were assumed in most of the work cited here to be pedestrians. Our work, however, should be equally applicable to any moving obstacles, including other robots and people in wheelchairs.

Several challenges are yet to be addressed by this model of the crowd. We fixed $\alpha$ at 1 to indicated that crowd behavior persisted and the robot did not forget, but smaller values would allow  the model to adapt to changing crowd behavior over time. Other considerations include doors that open and close, changes in the map, different shapes for robots and moving obstacles, and obstacles that vary their speed.

Thus far we have not considered how a nearby robot might cause people to move differently, although it known to have a significant effect on local trajectory prediction \cite{trautman2013robot}. CSA* does not consider that people might make way for the robot; in that case a simple A* plan might take less time than a CSA* plan. Another important factor is the spatial and temporal variance in the behavior of people around robots. For example, in a shopping mall, an area where seniors congregate should be treated differently from an area that attracts children, and areas that attract people from different cultures should also be treated differently. We intend to refine CSA* to learn how difficult it is to navigate in an area, rather than merely how crowded it is. Thus a crowded area where people give way to the robot would be treated differently from one where a similarly dense crowd is more obstructive.    
 
Crowd density maps are more broadly applicable than these experiments suggest. A museum-guide robot could use them to travel to the most crowded places, rather than away from them. A telepresence robot at a conference could use them to plan a path that allows the most interaction with conference attendees. Crowd density maps could also be used to guide active learning, that is, to direct the robot to areas where it can further observe movement (low $k_{ij}$ or $t_{ij}$ cells) to confirm low densities.

In summary, this paper presents a fast method that learns a crowd density map online, without the need for pedestrian trajectory datasets. This representation of global crowd behavior supports an agent's ability to generate crowd-sensitive plans. Our results demonstrate that crowd-sensitive planning reliably improves safety while it also reduces travel time and travel distance.   

\section{Acknowledgements}
This work was supported in part by NSF \#1625843.
\bibliographystyle{aaai}
\bibliography{references}

\end{document}